\documentclass[11pt,a4paper]{article}
\usepackage[hyperref]{acl2019}
\usepackage{times}
\usepackage{latexsym}
\usepackage{graphicx}
\usepackage{amsmath}
\usepackage{float}
\usepackage{booktabs}
\usepackage{url}
\usepackage{subcaption}
\makeatletter
\def\hlinewd#1{%
  \noalign{\ifnum0=`}\fi\hrule \@height #1 \futurelet
   \reserved@a\@xhline}
\makeatother

\aclfinalcopy % Uncomment this line for the final submission
\def\aclpaperid{2032} %  Enter the acl Paper ID here

\def\add{{\texttt{ADD}}}
\def\delete{{\texttt{DELETE}}}
\def\deletes{{\texttt{DELETE} }}
\def\keep{{\texttt{KEEP}}}
\def\keeps{{\texttt{KEEP} }}
\def\a{{\footnotesize \texttt{ADD}}}
\def\d{{\footnotesize \texttt{DEL} }}
\def\k{{\footnotesize \texttt{KEEP} }}
\def\del{\footnotesize \texttt{DEL}}
\def\seqlabel{Seq-Label}
\def\seqlabels{Seq-Label }

\def\wikilarge{WikiLarge}
\def\wikismall{WikiSmall}
\def\newsela{Newsela}
\def\editnet{EditNTS}

\title{EditNTS: An Neural Programmer-Interpreter Model for Sentence Simplification through Explicit Editing}

\author{Yue Dong$^1$, Zichao Li$^2$, Mehdi Rezagholizadeh$^2$, Jackie Chi Kit Cheung$^1$\\
$^1$MILA, McGill\\
{\texttt yue.dong2@mail.mcgill.ca, jcheung@cs.mcgill.ca}\\
$^2$Huawei Noah's Ark Lab\\
{\texttt \{li.zichao, mehdi.rezagholizadeh\}@huawei.com}}
\date{}

\begin{document}
\maketitle
\begin{abstract}
We present the first sentence simplification model that learns explicit edit operations (\add, \delete, and \keep) via a neural programmer-interpreter approach. Most current neural sentence simplification systems are variants of sequence-to-sequence models adopted from machine translation. These methods learn to simplify sentences as a byproduct of the fact that they are trained on complex-simple sentence pairs. By contrast, our neural programmer-interpreter is directly trained to predict explicit edit operations on targeted parts of the input sentence, resembling the way that humans might perform simplification and revision. Our model outperforms previous state-of-the-art neural sentence simplification models (without external knowledge) by large margins on three benchmark text simplification corpora in terms of SARI (+0.95 \wikilarge, +1.89 \wikismall, +1.41 \newsela), and is judged by humans to produce overall better and simpler output sentences\footnote{Link to our code and data can be found here \url{https://github.com/yuedongP/EditNTS}.}. 
\end{abstract}

\begin{table*}[h]
\resizebox{\linewidth}{!}{
\begin{tabular}{l|l}
\toprule
\multicolumn{2}{c}{\textbf{WikiLarge}} \\ \midrule
Source & \begin{tabular}[c]{@{}l@{}}\textcolor{blue}{in 2005 ,} meissner \textcolor{orange}{became} the second american woman to land the triple axel jump \textcolor{blue}{in national competition }.\end{tabular} \\ \hline
Output & meissner \textcolor{red}{was} the second american woman to land the triple axel jump . \\ \hline
Program & \textcolor{blue}{\d \d \d} \k  \textcolor{red}{\a(was)} \textcolor{blue}{\d} \k \k \k \k \k \k \k \k \k \k \textcolor{blue}{\d \d \d } \k \\ \hline
Reference & \begin{tabular}[c]{@{}l@{}}she is the second american woman and the sixth woman worldwide to do   a triple axel jump .\end{tabular} \\

\toprule
\multicolumn{2}{c}{\textbf{WikiSmall}} \\ \midrule
Source & \begin{tabular}[c]{@{}l@{}} \textcolor{blue}{theodoros `` thodoris ''} zagorakis -lrb- \textcolor{blue}{,} born october 27 ,  1971 \textcolor{blue}{in lyd -lrb- a village near the city} \\ \textcolor{blue}{of kavala -rrb- ,} is  a \textcolor{orange}{retired} greek \textcolor{orange}{footballer} \textcolor{blue}{and was the captain of the greece national  football team that} \\ \textcolor{blue}{won the 2004 uefa european football championship} . \end{tabular} \\ \hline
Output & zagorakis -lrb- born october 27 , 1971 is a \textcolor{red}{former} greek \textcolor{red}{football player} . \\ \hline
Program & \begin{tabular}[c]{@{}l@{}}\textcolor{blue}{\d \d \d \d} \k \k  \textcolor{blue}{\d} \k \k \k \k \k  \textcolor{blue}{\d \d \d \d \d \d \d \d \d \d } \\ \textcolor{blue}{\d \d}  \k \k  \textcolor{red}{\a(former)} \textcolor{blue}{\d} \k \textcolor{red}{\a(football) \a(player)} \textcolor{blue}{\d \d ... \d} \k \end{tabular} \\ \hline
Reference & theodoros zagorakis -lrb- born 27 october , 1971 -rrb- is a former football player .\\

\toprule
\multicolumn{2}{c}{\textbf{Newsela}} \\ \midrule
Source & \begin{tabular}[c]{@{}l@{}}schools and parent groups try to help \textcolor{orange}{reduce costs} for low-income students  \textcolor{blue}{who demonstrate a desire to}\\ \textcolor{blue}{play sports , she said} .\end{tabular} \\ \hline
Output & schools and parent groups try to help \textcolor{red}{pay} for low-income students . \\ \hline
Program & \begin{tabular}[c]{@{}l@{}}\k \k \k \k  \k \k \k \textcolor{red}{\a(pay)} \textcolor{blue}{\d \d} \k \k \k \textcolor{blue}{\d \d \d \d \d \d} \\  \textcolor{blue}{\d \d \d \d}  \k \end{tabular} \\ \hline 
Reference & \begin{tabular}[c]{@{}l@{}}clark said that schools do sometimes lower fees for students  who do n't have enough money .\end{tabular} \\ \bottomrule

\end{tabular}
}

\caption{Example outputs of \editnet{} taken from the validation set of three text simplification benchmarks. Given a complex source sentence, our trained model predicts a sequence of edit tokens (\editnet{} programs) that executes into a sequence of simplified tokens (\editnet{} output).}
\label{table:main_examples}
\end{table*}
\section{Introduction}

Sentence simplification aims to reduce the reading complexity of a sentence while preserving its meaning. Simplification systems can benefit populations with limited literacy skills \citep{watanabe2009facilita}, such as children, second language speakers and individuals with language impairments including dyslexia \citep{rello2013dyswebxia}, aphasia \citep{carroll1999simplifying} and autism \citep{evans2014evaluation}. 

Inspired by the success of machine translation, many text simplification (TS) systems treat sentence simplification as a monolingual translation task, in which complex-simple sentence pairs are presented to the models as source-target pairs \citep{zhang2017sentence}. Two major machine translation (MT) approaches are adapted into TS systems, each with its advantages:  statistical machine translation (SMT)-based models \citep{zhu2010monolingual,wubben2012sentence,narayan2014hybrid,xu2016optimizing} can easily integrate human-curated features into the model, while neural machine translation (NMT)-based models \citep{nisioi2017exploring,zhang2017sentence,vu2018sentence} can operate in an end-to-end fashion by extracting features automatically. Nevertheless, MT-based models must learn the simplifying operations that are embedded in the parallel complex-simple sentences \emph{implicitly}. These operations are relatively infrequent, as a large part of the original complex sentence usually remains unchanged in the simplification process
\citep{zhang2017constrained}. This leads to MT-based models that often produce outputs that are identical to the inputs \citep{zhao2018integrating}, which is also confirmed in our experiments. 

We instead propose a novel end-to-end Neural Programmer-Interpreter \citep{reed2016neural} that learns to \emph{explicitly} generate edit operations in a sequential fashion, resembling the way that a human editor might perform simplifications on sentences. Our proposed framework consists of  a programmer and an interpreter that operate alternately at each time step: the programmer predicts a simplifying edit operation (program) such as \add, \delete, or \keep; the interpreter executes the edit operation  while maintaining a context and an edit pointer to assist the programmer for further decisions. Table~\ref{table:main_examples} shows sample runs of our model. 

Intuitively, our model learns to skip words that do not need to be modified by predicting \keep, so it can focus on simplifying the parts that actually require changes. An analogy can be drawn to residual connections popular in deep neural architectures for image recognition, which give  models the flexibility to directly copy parameters from previous layers if they are not the focus of the visual signal \citep{he2016deep}. In addition, the edit operations generated by our model are easier to interpret than the black-box MT-based seq2seq systems: by looking at our model's generated programs, we can trace the simplification operations used to transform complex sentences to simple ones. Moreover, our model offers control over the ratio of simplification operations. By simply changing the loss weights on edit operations, our model can prioritize different simplification operations for different sentence simplification tasks (e.g., compression or lexical replacement). 

The idea of learning sentence simplification through edit operations was attempted by \citet{alva2017learning}. They were mainly focused on creating better-aligned simplification edit labels (``silver'' labels) and showed that a simple sequence labelling model (BiLSTM)  fails to predict these silver simplification labels. We speculate that the limited success of their proposed model is due to the facts that the model relies on an external system and assumes the edit operations are independent of each other. We address these two problems by 1) using variants of Levenshtein distances to create edit labels that do not require external tools to execute;  2) using an interpreter to execute the programs and summarize the partial output sequence immediately before making the next edit decision.  Our interpreter also acts as a language model to regularize the operations that would lead to ungrammatical outputs, as a  programmer alone will output edit labels with little consideration of context and grammar. In addition, our model is completely end-to-end and does not require any extra modules.%, in contrast to the requirement of external lexical replacement model LEXenstein \citep{paetzold2017lexical} for word replacement in \citet{alva2017learning}.

% 1) We propose a novel end-to-end Neural Programmer-Interpreter model for sentence simplification (\editnet{}): the programmer learns to predict simplification edits explicitly, while the interpreter executes these edits to generate the output and provides proper context to the programmer.   
% 2) Our model significantly outperforms state-of-the-art neural MT-based sentence simplification models by large margins in terms of SARI and judged by humans as fluent, simpler, and overall better.
% 3) The edit operations generated by our model are easier to interpret than the black-box MT-based seq2seq systems: by looking at our model's generated program traces, we can trace the simplification operations used to transform complex sentences to simple ones.
% 4) Our model offers control over the ratio of simplification operations. By simply changing the loss weights on edit operations, our model can prioritize different simplification operations for different sentence simplification tasks (e.g., compression or lexical replacement). 

Our contributions are two-fold:
1) we propose to model the edit operations \textit{explicitly} for sentence simplification in an end-to-end fashion, rather than relying on MT-based models to learn the simplification mappings implicitly, which often generates outputs by blindly repeating the source sentences; 2) we design an NPI-based model that simulates the editing process by a programmer and an interpreter, which outperforms the state-of-the-art neural MT-based TS models by large margins in terms of SARI and is judged by humans as simpler and overall better.

\section{Related Work}
%In this section, we review three areas of previous work that inspired our framework: 1) machine translation-based sentence simplification approaches, 2) edit-based sentence simplification methods, and 3) neural programmer-interpreter models.
\paragraph{MT-based Sentence Simplification}
SMT-based models and NMT-based models have been the main approaches for sentence simplification. They rely on learning simplification rewrites \textit{implicitly} from complex-simple sentence pairs. For SMT-based models, \citet{zhu2010monolingual}  adopt a tree-based SMT model for sentence simplification; \citet{woodsend2011learning} propose a quasi-synchronous grammar and use integer linear programming to score the simplification rules; \citet{wubben2012sentence} employ a phrase-based MT model to obtain candidates and re-rank them based on the dissimilarity to the complex sentence; \citet{narayan2014hybrid} develop a hybrid model that performs sentence splitting and deletion first and then re-rank the outputs similar to \citet{wubben2012sentence}; \citet{xu2016optimizing} propose SBMT-SARI, a syntax-based machine translation framework that uses an external knowledge base to encourage simplification. On the other side, many NMT-based models have also been proposed for sentence simplification: \citet{nisioi2017exploring} employ vanilla recurrent neural networks (RNNs) on text simplification; \citet{zhang2017sentence} propose to use reinforcement learning methods on RNNs to optimize a specific-designed reward based on simplicity, fluency and relevancy; \citet{vu2018sentence} incorporate memory-augmented neural networks for sentence simplification; \citet{zhao2018integrating} integrate the transformer architecture and PPDB rules to guide the simplification learning; \citet{sulem2018simple} combine neural MT models with sentence splitting modules for sentence simplification. 

\paragraph{Edit-based Sentence Simplification}
The only previous work on sentence simplification by \textit{explicitly} predicting simplification operations is by \citet{alva2017learning}. 
\citet{alva2017learning} use MASSAlign \citep{paetzold2017massalign} to obtain `silver' labels for simplification edits and employ a BiLSTM to sequentially predict three of their silver labels---\texttt{KEEP}, \texttt{REPLACE} and \texttt{DELETE}. 
%Although nine possible silver labels are proposed, only three of them can be predicted by their system due to the lack of good modules for applying the remaining operations. 
Essentially, their labelling model is a non-autoregressive classifier with three classes, where a downstream module \citep{paetzold2017lexical} is required for applying the \texttt{REPLACE} operation and providing the replacement word.  %to find 
% As a first attempt, their sequence labelling model reported limited success on correctly predicting the labels, but achieved comparable performance to the baseline NMT-based system \citep{nisioi2017exploring}. In contrast, our neural programmer-interpreter model is autoregressive and predicts $|V|+3$ operations, where $V$ is the set of words that can be added by our model and the rest three operations are \delete, \keep, and \texttt{STOP}. In other words, when our model predicts \add, it has to predict which word to add explicitly.  Model-wise, their BiLSTM-based sequence labelling model can be seen as a simplified non-autoregressive version of our programmer without the RNN interpreter, which has little consideration over the context nor the grammar for predicting edit operations.  
We instead propose an end-to-end neural programmer-interpreter model for sentence simplification,  which does not rely on external simplification rules nor alignment tools\footnote{Our model can be combined with these external knowledge base and alignment tools for further performance improvements.}.

\paragraph{Neural Programmer-Interpreter Models}
The neural programmer-interpreter (NPI) was first proposed by \citet{reed2016neural} as a machine learning model that learns to execute programs given their execution traces. Their experiments demonstrate success for 21 tasks including performing addition and bubble sort.  It was adopted by \citet{ling2017program} to solve algebraic word
problems and by \citet{berard2017lig, vu2018automatic} to perform automatic post-editing on machine translation outputs. 
%
%Our model's architecture falls into the category of NPI frameworks by generating programs before the actual outputs, but is different from the aforementioned models. \citet{reed2016neural} and \citet{ling2017program} use NPI to generate rationales by composing different parts of the inputs, where our model generating programs in a sequential fashion by revisiting the input complex sentence.
%. 
We instead design our NPI model to take monolingual complex input sentences and learn to perform simplification operations on them.

\begin{figure*}[h]
\centering
\includegraphics[width=0.99\textwidth]{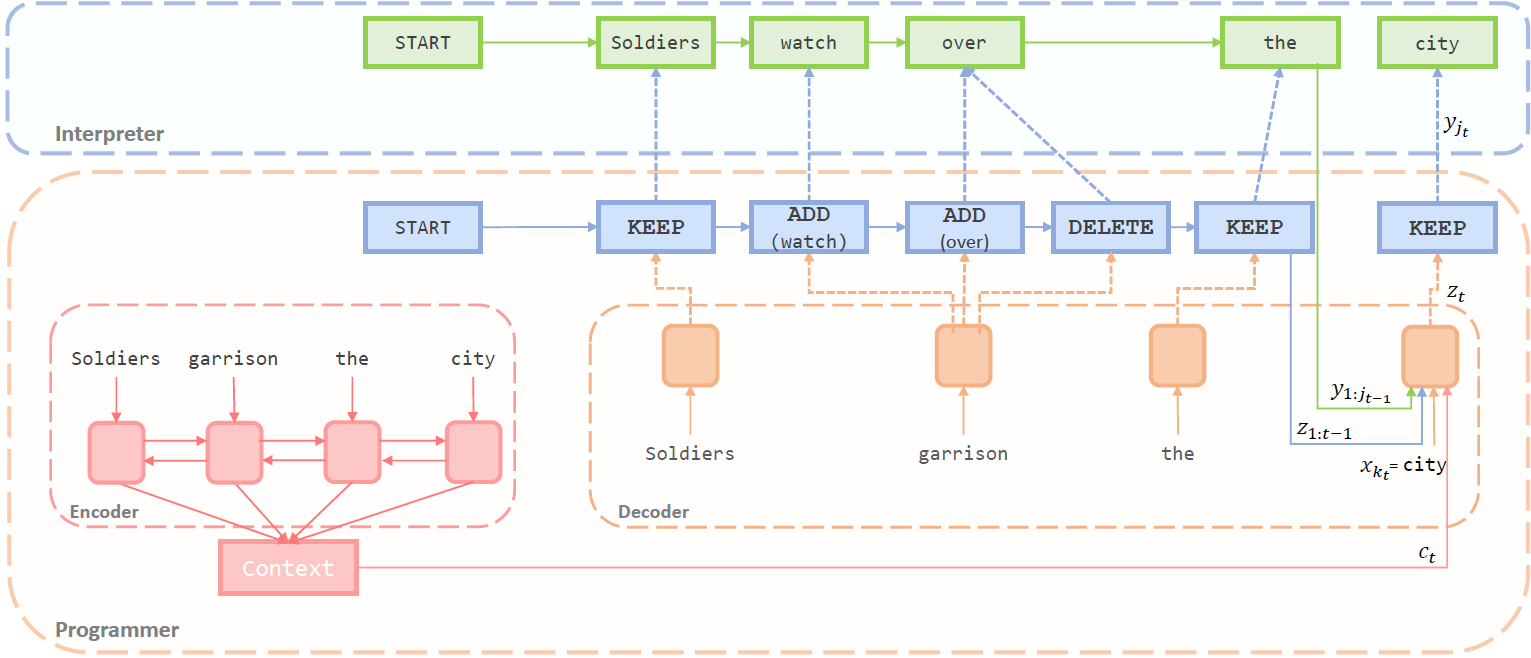}
\caption{Our model contains two parts: the programmer and the interpreter. At time step $t$, the programmer predicts an edit operation $z_t$ on the complex word $x_{k_t}$ by considering the interpreter-generated words $y_{1:j_{t-1}}$,  programmer-generated edit labels $z_{1:t-1}$, and a context vector $c_t$ obtained by attending over all words in the complex sentence. The interpreter executes the edit operation $z_t$  to generate the simplified token $y_{j_{t}}$  and provides the interpreter context $y_{1:j_{t}}$ to the programmer for the next decision. }
\label{fig:model}
\end{figure*}

\section{Model}
Conventional sequence-to-sequence learning models map a sequence $\mathbf{x}=x_1, \ldots, x_{\mathbf{|x|}} $ to another one $\mathbf{y}=y_1, \ldots, y_{|\mathbf{y}|}$, where elements of $\mathbf{x}$ and $\mathbf{y}$ are drawn from a vocabulary of size $V$, by modeling the conditional distribution $P(y_t|y_{1:t-1}, \mathbf{x})$ directly.
%However, models trained in this paradigm may fail to learn fine-grained edit operations and instead converge to the undesirable but safe solution of repeating the source sentence \cite{zhao2018integrating}, when the edit ratio from $\mathbf{x}$ to $\mathbf{y}$ is small.
Our proposed model, \editnet{}, tackles sentence simplification in a different paradigm by learning the simplification operations explicitly. An overview of our model is shown in Figure \ref{fig:model}.
\subsection{EditNTS Model}
\label{sec:editNTS}

\editnet{} frames the simplification process as executing a sequence of edit operations on complex tokens monotonically. We define the edit operations as $\{\text{\add(W), \keep, \delete, \texttt{STOP}}\}$.  Similar to the sequence-to-sequence learning models, we assume a fixed-sized vocabulary of $V$ words that can be added. Therefore, the number of prediction candidates of the programmer is $V+3$ after including \keep, \delete, and \texttt{STOP}. To solve the out-of-vocabulary (OOV) problem, conventional Seq2Seq models utilize a copy mechanism~\cite{gu2016incorporating} that selects a word from source (complex) sentence directly with a trainable pointer. In contrast, EditNTS has the ability to copy OOV words into the simplified sentences by directly learning to  predict \keep{} on them in complex sentences. We argue that our method has advantage over a copy mechanism in two ways: 1) our method does not need extra parameters for copying; 2) a copy mechanism may lead to the model copying blindly rather than performing simplifications.

We detail other constraints on the edit operations in Section~\ref{sec:label_creation}. It turns out that the sequence of edit operations $\mathbf{z}$ constructed by Section~\ref{sec:label_creation} is deterministic given $\mathbf{x}$ and $\mathbf{y}$  (an example of of $\mathbf{z}$ can be seen in Table~\ref{example}). Consequently, \editnet{} can learn to simplify by modelling the conditional distribution $P(\mathbf{z}|\mathbf{x})$ with a programmer, an interpreter and an edit pointer:
\begin{equation} 
\begin{split}\label{eq:decomposition}
P(\mathbf{z}|\mathbf{x}) & = \prod_{t=1}^{\mathbf{z}} P(z_t|y_{1:j_{t-1}},z_{1:t-1},x_{k_t}, \mathbf{x}).
\end{split}
\end{equation}

\begin{table}[h]
\footnotesize
\resizebox{\linewidth}{!}{
\begin{tabular}{l}
\hline
Complex sentence  $\mathbf{x}=x_1,\ldots x_{|\mathbf{x}|}$\\
{[}'the', 'line', 'between', 'combat', 'is', 'getting', 'blurry'{]} \\ \hline
Simple sentence  $\mathbf{y}=y_1,\ldots y_{|\mathbf{y}|}$\\
{[}'war', 'is', 'changing'{]} \\\hline
 Supervised programs  $\mathbf{z}=z_1,\ldots,z_{|\mathbf{z}|}$ \\
{[}\add('war'), \del, \del, \del, \del, 
\keep, \add('changing'), \\\del, \del{]} \\\hline
\end{tabular}
}
\caption{Given the source sentence $\mathbf{x}$ and the target sentence $\mathbf{y}$, our label creation algorithm (section \ref{sec:label_creation}) generates a deterministic program sequence  $\mathbf{z}$ for training.}
\label{example}
\end{table}

\noindent At time step $t$, the programmer decides an edit operation $z_t$ on the word $x_{k_t}$, which is assigned by the edit pointer,  based on the following contexts:  1) the summary of partially edited text $y_{1:j_{t-1}}$, 2) the previously generated edit operations $z_{1:t-1}$, 3) and the complex input sentence $\mathbf{x}$. The interpreter then executes the edit operation $z_t$ into a simplified token $y_{j_t}$ and updates the interpreter context based on $y_{1:j_{t}}$ to help the programmer at the next time step. The model is trained to maximize Equation \ref{eq:decomposition} where $\mathbf{z}$ is the expert edit sequence created in \ref{sec:label_creation}.  We detail the components and functions of the programmer and the interpreter hereafter.
  
\paragraph{Programmer.} The programmer employs an encoder-decoder structure to generate programs; i.e., sequences of edit operations $\mathbf{z}$. An encoder transforms the input sentence $\mathbf{x}=x_1,\ldots x_{|\mathbf{x}|}$ into a sequence of latent representations $h_i^{enc}$. We additionally utilize the part-of-speech (POS) tags $\mathbf{g}=g_1,\ldots g_{|\mathbf{x}|}$ to inject the syntactic information of sentences into the latent representations. The specific transformation process is:  
\begin{equation} \label{eq:h_enc}
    h^{enc}_i=\text{LSTM}^{enc}([e_1(x_i),e_2(g_i)])
\end{equation}
where $e_1(\cdot)$ and $e_2(\cdot)$ are both look-up tables.
The  decoder  is  trained  to  predict  the  next  edit label $z_t$ (Eq. \ref{eq:edit_pred}), given the vector representation $h^{enc}_{k_t}$  for the word $x_{k_t}$ that currently needs to be edited (Eq. \ref{eq:h_enc}), vector representation  $h^{\textit{edit}}_t$ of previously generated edit labels $z_{1:t-1}$ (Eq. \ref{eq:h_edit}), the  source context  vector $c_t$  (Eq.\ref{eq:context}), and the vector representation of previously generated words by the interpreter $y_{1:j_{t-1}}$ (Eq. \ref{eq:h_int}). 
\begin{align}
     P_{\textit{edit}}=\text{softmax}(V'(\text{tanh}(V(h^{\textit{edit}}_t)) \label{eq:edit_pred}\\
      h^{\textit{edit}}_t=\text{LSTM}^{\textit{edit}} ([h^{\textit{enc}}_{k_t},c_t,h^{\textit{edit}}_{t-1},h^{\textit{int}}_{t-1}]) \label{eq:h_edit} \\ 
     c_{t}=\sum_{j=1}^{|\mathbf{x}|}\alpha_{tj}h_j, \alpha_{tj}=\text{softmax}(h_{k_t},h_j) \label{eq:context}
\end{align}

Note that there are three attentions involved in the computation of the programmer. 1) the soft attention over all complex tokens to form a context $c_t$; 2) $k_{t}$:  the hard attention over complex input tokens for the \textbf{edit pointer}, which determines the index position of the current word that needs to be edited at $t$. We force $k_t$ to be  the number of \keeps and \deletes previously predicted by the programmer up to time $t$. 3) $j_{t-1}$: the hard attention over simple tokens for training (this attention is used to speed up the training), which is the number of \keeps and \add(W) in the reference gold labels up to time $t-1$. During inference, the model no longer needs this attention and instead incrementally obtains $y_{1:j_{t-1}}$ based on its predictions. 

\paragraph{Interpreter.} The interpreter contains two parts: 1) a parameter-free executor $\textit{exec}(z_t, x_{k_t})$ that applies the predicted edit operation $z_t$ on word $x_{k_t}$, resulting in a new word $y_{j_t}$. The specific execution rules for the operations are as follows: execute \keep/\deletes to keep/delete the word and move the edit pointer to the next word; execute \add(W) to add a new word W and the edit pointer stays on the same word; and execute \texttt{STOP} to terminate the edit process. 2) an LSTM interpreter (Eq.~\ref{eq:h_int}) that summarizes the partial output sequence of words produced by the executor so far. The output of the LSTM interpreter is given to the programmer in order to generate the next edit decision. 
\begin{equation}\label{eq:h_int}
    h^{int}_t=\text{LSTM}^{int} ([h^{int}_{t-1},y_{j_{t-1}}])
\end{equation}

\subsection{Edit Label Construction}
\label{sec:label_creation}
Unlike neural seq2seq models, our model requires expert programs for training. We construct these expert edit sequences from complex sentences to simple ones by computing the shortest edit paths using a dynamic programming algorithm similar to computing Levenshtein distances without substitutions. When multiple paths with the same edit distance exist, we further prioritizes the path that \texttt{ADD} before \delete. By doing so, we can generate a unique edit path from a complex sentence to a simple one, reducing the noise and variance that the model would face \footnote{We tried other way of labelling, such as 1) preferring \deletes to \texttt{ADD}; 2) deciding randomly when there is a tie; 3) including \texttt{REPLACE} as an operation. However, models trained with these labelling methods do not give good results from our empirical studies.}. Table~\ref{example} demonstrates an example of the created edit label path and Table~\ref{table:edit_labels} shows the counts of the created edit labels on the training sets of the three text simplification corpora.

\begin{table}[h]
\resizebox{\linewidth}{!}{
\begin{tabular}{l|l|l|l|l}
\toprule
 & \keep & \delete & \add & \texttt{STOP} \\ \midrule
\wikilarge & 2,781,648 & 3,847,848 & 2,082,184 & 246,768 \\
\wikismall & 1,356,170  & 780,482 & 399,826 & 88,028\\
\newsela & 1,042,640 & 1,401,331 & 439,110 & 94,208 \\ \bottomrule
\end{tabular}
}
\caption{Counts of the edit labels constructed by our label edits algorithm on three dataset (identical complex-simple sentence pairs are removed).}
\label{table:edit_labels}
\end{table}

As can be seen from Table~\ref{table:edit_labels}, our edit labels are very imbalanced, especially on \delete{}.  We resolve this by two approaches during training: 1) we associate the inverse of edit label frequencies as the weights to calculate the loss; 2) the model only executes \delete{} when there is an explicit \delete{} prediction. Thus, if the system outputs \texttt{STOP} before finish editing the whole complex sequence, our system will automatically pad \keep{} until the end of the sentence, ensuring the system outputs remain conservative with respect to the complex sequences.

\section{Experiments}
\label{sec:experiments}
\subsection{Dataset}
\label{sec:dataset}
Three benchmark text simplification datasets are used in our experiments. \textbf{WikiSmall} contains automatically aligned complex-simple sentence pairs from standard to simple English Wikipedia \citep{zhu2010monolingual}. We use the standard splits of 88,837/205/100 provided by \citet{zhang2017sentence} as train/dev/test sets. \textbf{WikiLarge} \citep{zhang2017sentence} is the largest TS corpus with 296,402/2000/359 complex-simple sentence pairs for training/validating/testing, constructed by merging previously created simplification corpora \cite{zhu2010monolingual,woodsend2011learning,kauchak2013improving}. In addition to the automatically aligned references, \citet{xu2016optimizing} created eight more human-written simplified references for each complex sentence in the development/test set of WikiLarge. The third dataset is \textbf{Newsela} \citep{xu2015problems}, which consists of 1130 news articles. Each article is rewritten by professional editors four times for children at different grade levels (0-4 from complex to simple). We use the standard splits provided by \citet{zhang2017sentence}, which contains 94,208/1129/1076 sentence pairs for train/dev/test. Table~\ref{table:dataset_stats} provides other statistics on these three benchmark training sets.
\begin{table}[h]
\small
\centering
\begin{tabular}{l|cc|cc}
\toprule
  & \multicolumn{2}{c|}{Vocabulary size} & \multicolumn{2}{c}{Sentence length} \\
 & comp & simp & comp & simp \\ \midrule
WikiLarge & 201,841 & 168,962 & 25.17 & 18.51 \\
WikiSmall & 113,368 & 93,835 & 24.26 & 20.33 \\
Newsela & 41,066 & 30,193 & 25.94 & 15.89 \\ \bottomrule
\end{tabular}
\caption{Statistics on the vocabulary sizes and the average sentence lengths of the complex and simplified sentences in the three text simplification training sets.}
\label{table:dataset_stats}
\end{table}

\subsection{Baselines}
We compare against three state-of-the-art SMT-based TS systems: PBMT-R \citep{wubben2012sentence} where the phrase-based MT system's outputs are re-ranked; 2) Hybrid \citep{narayan2014hybrid} where syntactic transformation such as sentence splits and deletions are performed before re-rank; 3) SBMT-SARI \citep{xu2016optimizing}, a syntax-based MT framework with external simplification rules. We also compare against four state-of-the-art NMT-based TS systems: vanilla RNN-based model NTS \cite{nisioi2017exploring}, memory-augmented neural networks $\text{N}_{\text{SE}}\text{L}_{\text{STM}}$ \cite{vu2018sentence}, deep reinforcement learning-based neural network DRESS and  DRESS-LS \citep{zhang2017sentence}, and DMASS+DCSS \citep{zhao2018integrating} that integrates the transformer model with external simplification rules. In addition, we compare our NPI-based \editnet{} with the BiLSTM sequence labelling model \citep{alva2017learning} that are trained on our edit labels\footnote{We made a good faith reimplementation of their model and trained it with our created edit labels. We cannot directly compare with their results because their model is not available and their results are not obtained from standard splits.}, we call it \seqlabels model. 
\subsection{Evaluation}
\label{sec:eval}
We report two widely used sentence simplification metrics in the literature: SARI \citep{xu2016optimizing} and FKGL \cite{kincaid1975derivation}. FKGL \citep{kincaid1975derivation} measures the readability of the system output (lower FKGL
implies simpler output) and SARI \citep{xu2016optimizing} evaluates the system output by comparing it against the source and reference sentences. Earlier work also used BLEU as a metric, but recent work has found that it does not reflect simplification \citep{xu2016optimizing} and is in fact negatively correlated with simplicity \citep{sulem2018bleu}. Systems with high BLEU scores are thus biased towards copying the complex sentence as a whole, while SARI avoids this by computing  the  arithmetic  mean  of the $N$-gram ($N \in \{1,2,3,4\}$) F1-scores of three rewrite operations: add, delete, and keep. We also report the F1-scores of these three operations. In addition, we report the percentage of unchanged sentences that are directly copied from the source sentences. We treat SARI as the most important measurement in our study, as \citet{xu2016optimizing} demonstrated that SARI has the highest correlation with human judgments in sentence simplification tasks. 

In addition to automatic evaluations, we also report human evaluations\footnote{The outputs of PBMT-R, Hybrid, SBMT-SARI and DRESS are publicly available and we are grateful to Sanqiang Zhao for providing their system's outputs.} of our system outputs compared to the best MT-based systems, external knowledge-based systems, and \seqlabels by three human judges\footnote{Three volunteers (one native English Speaker and two non-native fluent English speakers) are participated in our human evaluation, as one of the goal of our system is to make the text easier to understand for non-native English speakers. The volunteers are given complex setences and different system outputs in random order, and are asked to rate from one to five (the higher the better) in terms of simplicity, fluency, and adequacy.} with a five-point Likert scale. The volunteers are asked to rate simplifications on three dimensions: 1) fluency (is the output grammatical?), 2) adequacy (how much meaning from the original sentence is preserved?), and 3) simplicity (is the output simper than the original sentence?).

\subsection{Training Details}
We used the same hyperparameters across the three datasets. We initialized the word and edit operation embeddings with
100-dimensional GloVe vectors \citep{pennington2014glove} and the part-of-speech tag \footnote{We used the NLTK toolkit with the default Penn Treebank Tag set to obtain the part-of-speech tags; there are 45 possible POS-tags (36 standard tags and 7 special symbols) in total.} embeddings with 30 dimensions. The number of hidden units was set to 200 for the encoder, the edit LSTM, and the LSTM interpreter. During
training, we regularized the encoder with a dropout rate of 0.3 \citep{srivastava2014dropout}. For optimization, we used Adam \citep{kingma2014adam} with a learning rate 0.001 and weight decay of $10^{-6}$. The gradient was clipped to 1 \citep{pascanu2013difficulty}. We used a vocabulary size of 30K and the remaining words were replaced with UNK. In our main experiment, we used the inverse of the edit label frequencies as the loss weights, aiming to balance the classes. 
Batch size across all datasets was 64. 
%comment: xianzai xie dao zheli le keyi ba ta la chulai xiedao model de bufen
%\vspace{-3pt}
\section{Results}
\vspace{-3pt}
\begin{table}[!h]

\begin{subtable}[h]{\linewidth}
\resizebox{\linewidth}{!}{
\begin{tabular}[t]{l|l|lll|l|l}
\toprule
\textbf{WikiLarge} & SARI & \multicolumn{3}{l|}{Edit F1 of SARI} &FKGL&\% unc. \\
  & & add & del & keep & &\\ \midrule
 Reference & - & - & - & - &8.88& 15.88  \\ \midrule

\multicolumn{7}{l}{\textbf{MT-based TS Models}} \\
PBMT-R & \textbf{38.56} & \textbf{5.73} & 36.93 & 73.02 &8.33 & \textbf{10.58} \\
Hybrid & 31.40 & 1.84 & \textbf{45.48} & 46.87 &\textbf{4.57}&36.21 \\
NTS & 35.66 & 2.99 & 28.96 &  \textbf{75.02} &8.42&43.45\\
$\text{N}_{\text{SE}}\text{L}_{\text{STM}}$  & 36.88 & - & - && - - &\\
DRESS & 37.08 & 2.94& 43.15 & 65.15 &6.59&22.28\\
DRESS-LS & 37.27 & 2.81 & 42.22& 66.77 &6.62&27.02 \\
\midrule

\multicolumn{7}{l}{\textbf{Edit Labelling-based TS Models}} \\
\seqlabel& 37.08 & 2.94 & 43.20 &  65.10 &5.35&19.22\\
\editnet{} & 38.22 & 3.36 & 39.15 &  72.13 &7.30&10.86 \\
\midrule 
\midrule 
\multicolumn{7}{l}{\textbf{Models that use external knowledge base}} \\ 
SBMT-SARI& 39.96 & 5.96 & 41.42 & 72.52  &7.29&9.47 \\
DMASS+DCSS & 40.45 & 5.72 & 42.23 & 73.41 &7.79 &6.69\\\bottomrule
\end{tabular}
}
\caption{WikiLarge}
\end{subtable}

\vspace{5pt}

\begin{subtable}[h]{\linewidth}
\resizebox{\linewidth}{!}{
\begin{tabular}{l|l|lll|l|l}
\toprule
\textbf{WikiSmall} & SARI & \multicolumn{3}{l|}{Edit F1 of SARI} &FKGL&\% unc. \\ 
   && add & del & keep  && \\ \midrule
 Reference & - & - & - &- &8.86&  3.00\\ \midrule
 \multicolumn{7}{l}{\textbf{MT-based TS Models}} \\
PBMT-R  & 15.97 & 6.75 & 28.50 & 12.67 &11.42&14.00 \\
Hybrid & 30.46 &\textbf{16.53}& 59.60 & \textbf{15.25} &9.20&4.00 \\
NTS & 13.61 & 2.08 & 26.21 & 12.53&11.35 &36.00 \\
$\text{N}_{\text{SE}}\text{L}_{\text{STM}}$ & 29.75 & - & - & - & -&-\\ 
DRESS & 27.48 & 2.86 & 65.94 & 13.64 &7.48 &11.00 \\
DRESS-LS & 27.24 & 3.75&64.27 & 13.71  &7.55&13.00 \\ \midrule
\multicolumn{7}{l}{\textbf{Edit Labelling-based TS Models}}\\
\seqlabel  & 30.50 & 2.72 & 76.31 & 12.46 &9.38&9.00\\
\editnet{}   & \textbf{32.35} & 2.24 & \textbf{81.30} & 13.54 &\textbf{5.47} &\textbf{0.00}  \\ \bottomrule
\end{tabular}
}
\caption{WikiSmall}
\end{subtable}

\vspace{5pt}

\begin{subtable}[h]{\linewidth}
\resizebox{\linewidth}{!}{
\begin{tabular}{l|l|lll|l|l}
\toprule
\textbf{Newsela}  & SARI & \multicolumn{3}{l|}{Edit F1 of SARI} & FKGL&\%unc. \\
  & &add & delete &keep &&  \\ \midrule
 Reference& - & - & - &-  &3.20& 0.00 \\ \midrule
 \multicolumn{7}{l}{\textbf{MT-based TS Models}} \\
PBMT-R & 15.77 & 3.07 & 38.34 & 5.90 &7.59&5.85  \\
Hybrid& 30.00 & 1.16 & 83.23 & 5.62  &4.01&\textbf{3.34} \\
NTS  & 24.12 & 2.73 & 62.66 & 6.98 &5.11&16.25 \\
$\text{N}_{\text{SE}}\text{L}_{\text{STM}}$ & 29.58 & - & - & -& -&- \\
DRESS & 27.37 & 3.08 & 71.61 & \textbf{7.43} &4.11&11.98 \\
DRESS-LS & 26.63 & \textbf{3.21} & 69.28 & 7.40 &4.20&15.51 \\ \midrule
\multicolumn{7}{l}{\textbf{Edit Labelling-based TS Models}} \\
\seqlabel& 29.53 & 1.40 & 80.25 & 6.94  &5.45&15.97 \\
\editnet{}  & \textbf{31.41} &1.84 & \textbf{85.36} & 7.04 &\textbf{3.40}&4.27 \\ \bottomrule
\end{tabular}
}
\caption{Newsela}
\end{subtable}
\caption{Automatic Evaluation Results on three benchmarks. We report corpus level FKGL, SARI and edit F1 scores (add,keep,delete). In addition, we report the percentage of unchanged sentences (\%unc.) in the system outputs when compared to the source sentences. }
\label{table:main_results}
\end{table}

\begin{table*}[ht]
\centering
\small
\begin{tabular}{l|llll|llll|llll}
\toprule
& \multicolumn{4}{c|}{\textbf{\wikilarge{}}} & \multicolumn{4}{c|}{\textbf{\newsela{}}} & \multicolumn{4}{c}{\textbf{\wikismall}} \\
 & F & A & S & avg. & F & A & S & avg. & F & A & S & avg. \\ \midrule
Reference & 4.39 & 4.11 & 2.62 & 3.71 &4.40  &2.74  & 3.79 & 3.64   &4.48 & 4.03  &2.99 & 3.83\\ \midrule
PBMT-R & 4.38 & 4.05 & 2.28 & 3.57 & 3.76 & \textbf{3.44} & 2.28 & 3.16 & 4.32 & \textbf{4.28} & 1.53 & 3.38 \\
Hybrid & 3.41 & 3.01 & \textbf{3.31} & 3.24 & 3.62& 2.88  & 2.97 & 3.16 &  3.76 & 3.87 & 2.12 & 3.25\\
SBMT-SARI & 4.25 & 3.96 & 2.61 & 3.61 & - & - &-  & - & - & - & - & - \\ 
DRESS & 4.63 & 4.01 & 3.07 & 3.90 & 4.16 & 3.08 &3.00  & 3.41 & \textbf{4.61}  & 3.64 & 3.62 &3.96\\
%DRESS\_LS &  &  &  &  &  &  &  &  &  &  &  &  \\ \hline
DMASS+DCSS & 4.39 & 3.97 & 2.80 & 3.72 & - & - & - & - & - &-  &  -& - \\ 
seq-label & 3.91 & 4.11 & 2.97 & 3.66 & 3.45 & 3.22 & 2.09 & 2.92 & 3.83 &  3.9& 2.01 & 3.25 \\
\editnet{} & \textbf{4.76} & \textbf{4.45} & 3.18 & \textbf{4.13} &\textbf{4.34}  & 3.13 & \textbf{3.16} &  \textbf{ 3.54} &4.31 & 3.34  &\textbf{4.26} & \textbf{3.97}\\ \bottomrule
\end{tabular}
\caption{Mean ratings for Fluency (F), Adequacy (A), Simplicity (S), and the Average score (avg.) by human judges on the three benchmark test sets. 50 sentences are rated on \wikilarge, 30 sentences are rated on WikiSmall and \newsela. Aside from   comparing   system   outputs, we also include human ratings for the gold standard reference as an upper bound.}
\label{tab:human_eval}
\end{table*}

Table \ref{table:main_results} summarizes the results
of our automatic evaluations. In terms of readability, our system obtains lower (= better) FKGL compared to other MT-based systems, which indicates our system's output is easier to understand.  In terms of the percentage of unchanged sentences, one can see that MT-based models have much higher rates of unchanged sentences than the reference. Thus, the models learned a safe but undesirable strategy of copying the sources sentences directly. By contrast, our model learns to edit the sentences and has a lower rate of keeping the source sentences unchanged.

In term of SARI, the edit labelling-based models \seqlabels and \editnet{} achieve better or comparable results with respect to state-of-the-art MT-based models, demonstrating the promise of learning edit labels for text simplification. Compared to \seqlabel, our model achieves a large improvement of (+1.14,+1.85,+1.88 SARI) on \wikilarge, \newsela, and \wikismall. We believe this improvement is mainly from the interpreter in \editnet, as it provides the proper context to the programmer for making edit decisions (more ablation studies in section \ref{sec:ablation}). On Newsela and \wikismall{}, our model significantly outperforms  state-of-the-art TS models by a large margin (+1.89, +1.41 SARI), showing that  \editnet{} learns simplification better on smaller datasets with respect to MT-based simplification models. On \wikilarge{}, our model outperforms  the best NMT-based system DRESS-LS by a large margin of +0.95 SARI and achieves comparable performance to the best SMT-based model PBMT-R. While the overall SARI are similar between \editnet{} and PBMT-R,  the two models prefer different strategies: \editnet{} performs extensive \delete{} while PBMT-R is in favour of performing lexical substitution and simplification. 

On \wikilarge, two models SBMT-SARI and DMASS+DCSS reported higher SARI scores as they employ external knowledge base PPDB for word replacement. These external rules can provide reliable
guidance about which words to modify, resulting in higher add/keep F1 scores (Table~\ref{table:main_results}-a). On the contrary, our model is inclined to generate shorter sentences, which leads to high F1 scores on delete operations \footnote{As the full outputs of $\text{N}_{\text{SE}}\text{L}_{\text{STM}}$ are not available, we cannot compute the edit F1 scores and FKGL for this system.}. Nevertheless, our model is preferred by human judges than SBMT-SARI and DMASS+DCSS in terms of all the measurements (Table \ref{tab:human_eval}), indicating the effectiveness of our model on correctly performing deleting operations while maintaining fluent and adequate outputs.  Moreover, our model can be easily integrated with these external PPTB simplification rules for word replacement by adding a new edit label ``replacement'' for further improvements.

The results of our human evaluations are presented in Table~\ref{tab:human_eval}. As can be seen, our model outperforms MT-based models on Fluency, Simplicity, and Average overall ratings. Despite our system \editnet{} is inclined to perform more delete operations, human judges rate our system as adequate. In addition, our model performs significantly better than \seqlabels in terms of Fluency, indicating the importance of adding an interpreter to 1) summarize the partial edited outputs and 2) regularize the programmer as a language model. %As a consequence of having the Interpreter, our \editnet generates very fluent and grammatical correct sentences that are even better than NMT sequence to sequence models. 
Interestingly, similar to the human evaluation results in \citet{zhang2017sentence}, judges often prefer system outputs than the gold references.

\paragraph{Controllable Generation:}
In addition to the state-of-the-art performance, \editnet{} has the flexibility to prioritize different edit operations.  Note that NMT-based systems do not have this feature at all, as the sentence length of their systems' output is not controllable and are purely depends on the training data.  Table \ref{table:ablation2} shows that by simply changing the loss weights on different edit labels, we can control the length of system's outputs, how much words it copies from the original sentences and how much novel words the system adds.  
\begin{table}[h]
\resizebox{\linewidth}{!}{
\begin{tabular}{l|c|c|c}
\toprule
add:keep:delete ratio & Avg. len & \% copied & \% novel \\ \midrule
 10:1:1 (add rewarded)& 25.21 & 53.52&  56.28 \\
 1:10:1 (keep rewarded)& 21.52 & 84.22 & 12.81   \\
 1:1:10 (delete rewarded)& 15.83& 57.36 & 16.72  \\ \bottomrule
\end{tabular}
}
\caption{Results on Newsela by controlling the edit label ratios. We increase the loss weight on \add,\keep,\deletes ten times respectively. The three rows show the systems' output statistics  on the average output sentence length (Avg. len), the average percentage of tokens that are copied from the input (\% copied), and the average percentage of novel tokens that are added with respect to the input sentence (\% novel).  }
\label{table:ablation2}
\end{table} 

\subsection{Ablation Studies} 
\label{sec:ablation}
In the ablation studies, we aim to investigate the effectiveness of each component in our model. We compare the full model with its variants where POS tags removed, interpreter removed, context removed. As shown in Table~\ref{table:ablation1}, the interpreter is a critical part to guarantee the performance of the sequence-labelling model, while POS tags and attention  provide further performance gains. %In addition, human study have shown that having an interpreter will help the model to generate more fluent and adequate outputs, which are consequences of interpreter serving as a language model to regularize the edit operations. 
 \begin{table}[h] 
 \centering
\small 
\begin{tabular}{l|l|lll}
\toprule
Newsela & SARI & \multicolumn{3}{l}{Edit F1 of SARI}  \\
 &  & add & delete & keep   \\ \midrule
\editnet{} & 31.41 &1.84 & 85.36 & 7.04  \\
~~ $-$ POS tags  & 31.27 & 1.46 & 85.34 & 7.00 \\
~~ $-$ attn-context & 30.95 & 1.54 & 84.26 & 7.05 \\
~~ $-$ Interpreter  & 30.13 & 1.70 & 81.70 & 7.01  \\
%Seq-label & 29.53 & 1.40 & 80.25 & 6.94  \\ 
\bottomrule
\end{tabular}
\caption{Performance on Newsela after removing different components in \editnet.}
\label{table:ablation1}
\end{table}

\section{Conclusion}

We propose an NPI-based model for sentence simplification, where edit-labels are predicted by the programmer and then executed into simplified tokens by the interpreter. Our model outperforms previous state-of-the-art machine translation-based TS models in most of the automatic evaluation metrics and human ratings, demonstrating the effectiveness of learning edit operations \textit{explicitly} for sentence simplification. Compared to the black-box MT-based systems, our model is more interpretable by providing generated edit operation traces, and more controllable with the ability to prioritize different simplification operations. 
%In addition, our model has the flexibility of controlling the ratio In the future, we would like to incorporate syntactic simplification, such as sentence splitting into our \editnet framework. Another direction we believe will improve our model's performance is to incorporate external simplification rules, such as PPTB, to encourage more word replacement. 

\section*{Acknowledgments}
The research was supported in part by Huawei Noah's Ark Lab (Montreal Research Centre), Natural Sciences and Engineering Research Council of Canada (NSERC) and Canadian Institute For Advanced Research (CIFAR). We thank Sanqiang Zhao and Xin Jiang for sharing their pearls of wisdom, Xingxing Zhang for providing the datasets and three anonymous reviewers for giving  their insights and comments. %We would also express our appreciation to our sponsors, Huawei Noah's Ark Lab and NSERC for supporting Yue Dong and a Canada CIFAR AI Chair for supporting Jackie CK Cheung.

\bibliography{acl2019}
\bibliographystyle{acl_natbib}

\appendix
\end{document}